# Free-Flying Crew Cooperative Robots on the ISS:
## A Joint Review of Astrobee, CIMON, and Int-Ball Operations

Seiko Piotr Yamaguchi*[1], Andres Mora Vargas*[2], Till Eisenberg*[4], Christian Rogon[3],
Tatsuya Yamamoto[1], Shona Inoue[1], Christoph Kössl [4], Brian Coltin[2], Trey Smith[2], Jose V. Benavides[2]
*These Authors Contributed Equally

*Abstract—*. Intra-vehicular free-flying robots are anticipated to support various work in human spaceflight while working side-by-side with astronauts. Such example of robots includes NASA's Astrobee, DLR's CIMON, and JAXA's Int-Ball, which are deployed on the International Space Station. This paper presents the first joint analyses of these robot's shared experiences, co-authored by their development and operation team members. Despite the different origins and design philosophies, the development and operations of these platforms encountered various convergences. Hence, this paper presents a detailed overview of these robots, presenting their objectives, design, and onboard operations. Hence, joint lessons learned across the lifecycle are presented, from design to on-orbit operations. These lessons learned are anticipated to serve for future development and research as design recommendations.

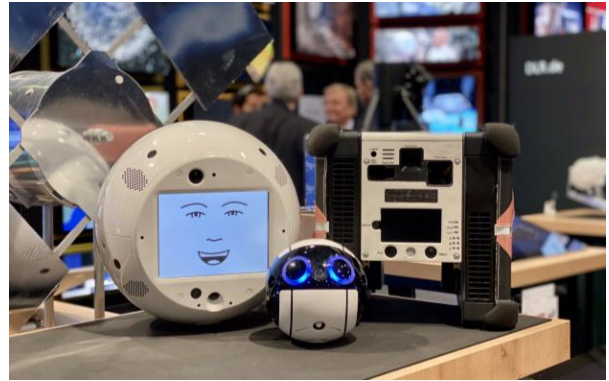

Fig. 1: ISS Free-Flying Robots (ground models).
Left: DLR's CIMON, center: JAXA's Int-Ball,
right: NASA's Astrobee ©JAXA/NASA/DLR

## I. INTRODUCTION

The vision of intelligent, autonomous robots working alongside humans in space stations has been anticipated to enhance human spaceflight operations. Such crew-assistive robots can enhance astronauts' operations, offload tedious or hazardous tasks, and enable new avenues of scientific research. In recent years, this vision has begun to materialize aboard the International Space Station (ISS), which serves as an unparalleled long-duration testbed for technologies and operational concepts required for sustained human-robot teaming in microgravity.

Free-flying robots are anticipated to work side-by-side with astronauts, dynamically moving around to improve situational awareness and improve ground-space coordination. ISS has been providing valuable experience with such free-flying robots from NASA's SPHERES experiments (2006~), and their successors: Astrobee robots (2019~), JAXA's Int-Ball (2017), and improved Int-Ball2 (2023-), as well as DLR (German Space Center)'s CIMON (2018~).

1: S. P. Yamaguchi, T. Yamamoto and S. Inoue are with Human Spaceflight Technology Directorate, Japan Aerospace Exploration Agency (JAXA), 2-1-1 Sengen, Tsukuba, Japan, 305-8505. e-mail: yamaguchi.seiko[at]jaxa.jp
2: A. Mora Vargas, B. Coltin, T. Smith, and J.V. Benavides are with Astrobee Facilities Team, NASA Ames Research Center, e-mail: andres.moravargas[at]nasa.gov
3: C. Rogon is with German Aerospace Center (DLR), Königswinterer Str. 522-524, 53227 Bonn, Germany, email: Christian.Rogon[at] dlr.de
4: T. Eisenberg and C. Köessl are with Airbus Defense and Space, 88039, Friedrichshafen, Baden-Württemberg, Germany, email: till.eisenberg [at]airbus.com

Each project had a distinct primary objective. Astrobee was designed as a versatile, open-source research platform intended to support a wide array of guest science and technology demonstrations. CIMON was a pioneering effort in human-machine interaction, using artificial intelligence to provide verbal assistance to the crew. The Int-Ball series was conceived as a dedicated, crew-assistive camera platform developed through an iterative design process.

Despite their different origins and design philosophies, the development and operations teams for these three platforms encountered remarkable convergences. This paper, authored collectively by members of all three teams, provides the first consolidated analysis of these shared experiences. Its objective is to synthesize the cross-platform lessons learned across the full project lifecycle to provide a foundational body of knowledge that can inform, de-risk, and accelerate the development of future free-flying robotic systems.

In this paper, we provide a detailed overview of these robot systems, covering their design, development history, and operational experience. Then, consolidated analysis of our joint lessons learned are presented, organized into the key areas of development, operations, and user integration. Finally, we conclude with a summary of our key findings and a perspective on the future of human-robot teaming in space.

## II. OVERVIEW OF THE ISS FREE-FLYING ROBOTS

This chapter introduces the overview of the free-flying robots on the ISS, including their background, design, and operational experience. Comparison of these robots is presented in Tab. 1.

Tab.1 : Specifications of free-flying robots deployed on the ISS.

| | Astrobee | CIMON-1 | CIMON-2 | Int-Ball | Int-Ball2 |
|---|---|---|---|---|---|
| **Objective** | General purpose science and payload facility | Enhanced AI crew companion | | Tech demo for remote controlled camera | Substitute crew work for imagery tasks |
| **Developed by:** | NASA | DLR | | JAXA | |
| **Deployed on:** | 2019 - present | 2018 | 2019 - present | 2017 - 2019 | 2023 - present |
| **Size / Shape** | 30 cm cube | 32cm sphere | | 15cm sphere | 20cm sphere |
| **Mass** | Approx. 10kg | Approx. 5kg | | Approx. 1kg | Approx. 3kg |
| **Mobility** | 12 nozzles with centrifugal impellers (300mN max. x-nozzle) | 12 propelling fans (about 120mN max x-direction) | | 12 fans (~3mN each) & 3reaction wheels | 8 propelling fans (60-163mN each fan) |
| **Navigation Method** | Graph-localizer | Stereo camera VSLAM + IMU sensor fusion | | Monocular Camera & Marker based | Stereo Visual-Inertial SLAM (feature-based SLAM + IMU sensor fusion) |
| **Imaging Capability** | 6 different cameras | ~Full HD (1280x1024@25fps) | | ~Full-HD | ~4K (1920×1080@30fps, 4056x3040@15fps) |
| **Crew Interface** | Touch screen, 2 flashlights, LED lights, perching arm, Reset/On-Off/Emergency buttons, Ground data System | LCD monitor, speaker, status LEDs, laser-pointer, on-off and mic-mute buttons , physical contact detection (IMU based) | | Manual on/off switch, "Eye" LEDs to indicate robot and imagery states. Physical contact detection (IMU based) | |
| **Sensors** | HazCam, Scicam, NavCam, DockCam, PerchCam, SpeedCam (IR), IMU, microphone | Crew camera, IR camera, IMU, ultrasonic sensors, temperature sensors | | Front camera, nav. side monocular camera, IMU, ultrasonic sensors, microphone | Front camera, Navigation stereo camera, IMU, microphone |
| **Computation architecture** | Tiered on-board (LLP/MLP/HLP) | Two units of on-board processors + ground AI server | | Single on-board processor | Single on-board processor |
| **Processors** | Wandboard Dual, InForce IFC6501, InForce IFC6601 | Pokini F WIFI (AMD A4-6700T) | | Armadillo-810 | Jetson TX2 |
| **Software** | Ubuntu/ ROS (LLP/MLP) Android (LLP) | Ubuntu/ROS | | Armadillo Base OS | Ubuntu/ROS |
| **Power System** | Li-Ion – Automatic docking & recharge | Li-Ion – replaceable by crew / wired power input | | Li-Ion – USB charge by crew | Li-Ion – Automatic docking & recharge |
| **Operation time (on batteries)** | 3h | 2h | 3h | 2h | 3h |
| **Data Communication** | Wireless-LAN(WLAN) / Wired-LAN over DS | Wired connection / WLAN / Bluetooth | | WLAN | WLAN/ UART to DS connected to laptop |
| **Payload / extension interface** | 3 payload bays, each with 1xUSB, power to payload | USB x2, Bluetooth | | N/A | Micro-USB + side attachment |
| **Ground Operations** | Ground Data System | Ground Station remote connection + AI Server | | Remote control with simple UI | Remote control with GUI |
| **Ground Verification / testing method** | Gazebo Simulator + Granite Table testing facility with two ground models | Software Simulations Ground Model testing Elegant Bread-Board (eBB), HIL simulation | | Granite table test with ground model (GM) | Gazebo Simulator, Granite Table with GM, H/W-in-the-loop sim. |

*A. Astrobee (NASA); A Modular Research Platform*

Since 2006, NASA's Synchronize, Position, Hold, Engage, Reorient, Experimental Satellites (SPHERES) supported guest science research on the ISS and became one of the most-used ISS payload facilities [1]. NASA's three Astrobee (Bumble, Honey, and Queen) became the next generation of free-flying robots on board of the ISS in 2019 replacing SPHERES [2].

Astrobee improved the capabilities SPHERES offered to the community by increasing autonomy and therefore reducing the amount of crew time required for experimental setup or regular operation; it does not depend on additional infrastructure on the station to navigate as it has a vision-based navigation system; it has been able to provide Guest Scientists with up to 3 hours of continuous operation and enables them to attach different types of payloads thanks to its three payload bays, such as Astrobee's perching arm [3]. Astrobee supports multiple research and technology demonstration efforts including manipulation, computer vision, Artificial Intelligence (AI), and human-robot interaction (HRI) [4, 5]. Astrobee and its Guest Scientists have produced several peer-reviewed conference and journal papers. Moreover, its source code, data from on-orbit activities and localization datasets are publicly released [6] as an effort to push forward the development of autonomous robots in microgravity environments.

The Astrobees are located at the Japanese Experimental Module (JEM) where the robots can recharge their Lithium-Ion batteries at the docking station [7, 8]. Each Astrobee measures approximately 30x30x30cm and weighs about 10 kg. Each of them propels itself through two battery-powered fans and twelve adjustable flow-rate nozzles. It has six different cameras that enable the robot to navigate inside the ISS (NavCam, SpeedCam, HazCam), perch onto handrails (PerchCam), provide high-definition video for Guest Scientists experiments (SciCam) and autonomously dock and undock (DockCam) from the docking station.

HRI research is enabled through its front and aft flashlights, laser pointer, LED lights that can be used to generate different lighting patterns, and a touch screen in its front face, which displays a looped video of "eyes" conveying the working state of the robot to crew.

Astrobee has three main computers, namely the Low-Level Processor (LLP), Mid-Level Processor (MLP), and the High-Level Processor (HLP). The LLP reliably runs high-rate control code in isolation from other software. The MLP oversees running most of the Astrobee Software including computer-vision based path planning. Finally, the main function of the HLP is to run Guest Science data and managing SciCam video compression, touch screen, and other less critical devices.

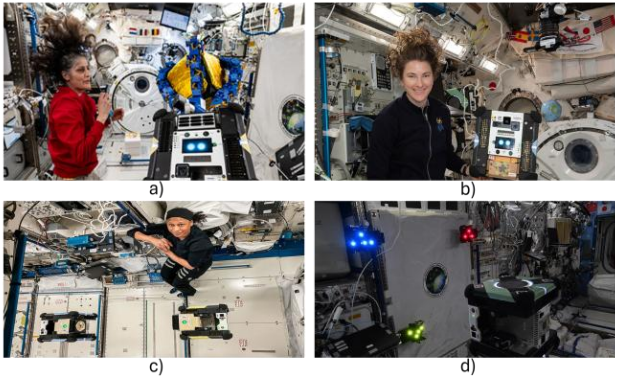

Fig. 2: Astrobee on-orbit activities at the ISS. Four different experiments: a) KMI's REACCH. b) Bosch USA/Astrobotics's SoundSee. c) University of Southern California's CLINGERS. d) Florida Institute of Technology's SVGS. Image credit: NASA

The Astrobee Robot Software (ARS) is available open source [9], where some of its key tasks are to localize throughout the JEM without extra infrastructure, precisely plan and execute motions without collision, provide control and monitoring, resilient ground control despite communication loss, and support for multiple control modes—remote teleoperation, autonomous execution, and on-board Guest Science software. ARS also handles autonomous docking for recharging and wired communication, energy-saving perching with pan/tilt camera use, and management of Guest Science software, hardware payloads, UI components, simulation, and data replay for testing. The LLP and MLP run Ubuntu (currently 20.04) because of its widespread use and the availability of software packages, notably Robot Operating System (ROS) [10]. The HLP, runs Android (Nougat 7.1) because it is the only OS supporting some key hardware for Astrobee (SciCam, video encoder and touchscreen). Android allows for the encapsulation of Guest Science software for the HLP as Android Packages (APKs), avoiding custom deployment and management methods.

As of May 2025, the Astrobee Facility has supported 17 unique Guest Scientists in 22 different projects, and is planned to support four additional Guest Scientists and seven additional projects between 2024 and 2025. Guest Scientists have been from academia (Stanford, MIT), industry (KMI, Obruta, Bosch USA, and more), or government (NASA, JAXA, CSIRO). Figure 2 shows some examples of Guest Scientists' in-orbit activities. Academic Guest Scientists have carried out research in autonomous spacecraft rendezvous and docking [11], mapping and navigation [12], detecting changes in the ISS [13], and semantic mapping and localization [7]. Astrobee has enable academic partners to demonstrate technologies for the first time in space robotics, for instance, the US Naval Research Laboratory's (NRL's) Autonomous Planning In-space Assembly Reinforcement-learning free-flYer (APIARY) team conducted the first ever reinforcement learning control of a free-flyer in space within a 3-month timeline.

Guest Scientists from the industry have successfully demonstrated their technologies leveraging Astrobee capabilities, some examples include Bosch's acoustic maps, KMI's on-orbit servicing, assembly, and manufacturing (OSAM) using its large gripper to capture a tumbling target, or Canadian corporation Obruta which builds software for spacecraft rendezvous, proximity operations, and docking (RPOD) to increase the sustainability of the space industry through in-space servicing such as refueling, life extension, or debris removal [14]. *Honey* autonomously docked with *Bumble* using Obruta's AI-based monocular vision-based software along with its custom guidance and control software.

With international space agency partners such as JAXA, Astrobee has been fundamental in its educational outreach role. Since 2019 JAXA and NASA have collaborated to carry out JAXA's Kibo Robotic Programming Challenge (Kibo-RPC). This international competition is intended for undergraduate university students in 2024 alone, with over 2760 students across the Asia Pacific region. Kibo-RPC is a programming challenge where students from around the world participate in teams and after passing a series of simulation-based qualifying rounds at their countries, each country selects one team to participate in the final event where students' code controls an Astrobee robot to solve a challenge in real-time on board of the ISS with crew participation. Similarly, the Massachusetts Institute of Technology (MIT) has led the Zero Robotics competition since 2022, where USA middle and high school students develop their own code to resolve a problem using one Astrobee. This competition has impacted over 700 students in 2024 alone.

Astrobee has benefited from not being a finished product but rather an extensible research platform that enables innovation across different domains. Thanks to the Guest Scientists and their varied investigations, the Astrobee team has incrementally grown the capabilities and features the Astrobee platform offers thereby furthering the innovation Guest Scientists can create.

### B. CIMON (DLR): An AI-Powered Crew Companion

As a pioneering technology demonstration, the Crew Interactive MObile companioN (CIMON) project was commissioned by the German Space Agency at DLR to explore future human-robot interaction on long-duration missions [15]. CIMON is an interactive AI-based assistant designed to reduce astronaut workload and stress, provide hands-free access to procedures and data, and serve as a research subject for studying human-AI teaming. Unlike task-oriented robots, CIMON's core function is to interact, converse, and demonstrate a form of "emotional intelligence," making it a unique presence on the ISS [16].

CIMON was developed by Airbus, and first launched in 2018, located and operated within the European Columbus module. CIMON unit is a sphere with a diameter of 32 cm and a mass of approximately 5 kg. It can propel itself using 14 internal, battery-powered fans that provide attitude control and enable navigation within the module. The robot's primary interface is a front-facing LCD screen that displays an animated "face," conveying operational status and facilitating a more natural interaction. It is equipped with several cameras, including stereo cameras for 3D vision and navigation, an infrared camera, two side cameras to support augmented reality tasks, and a high-resolution camera for documenting experiments. Its audio system uses multiple microphones to perform directional sound source localization, enabling it to turn and face the astronaut who is speaking.

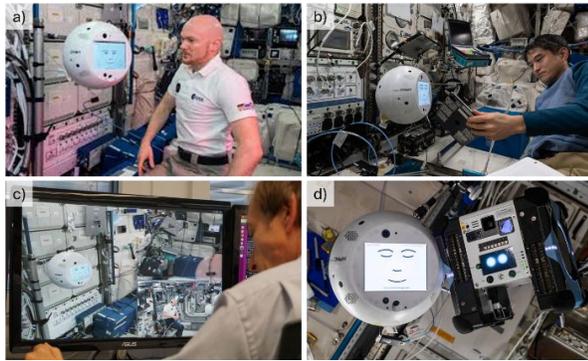

Fig. 3: DLR's CIMON. a) CIMON with ESA astronaut Alexander Gerst, b) CIMON being prepared for a task by JAXA astronaut Takuya Onishi, c) CIMON ground control, d) CIMON & Astrobee. Image credit: DLR/ESA

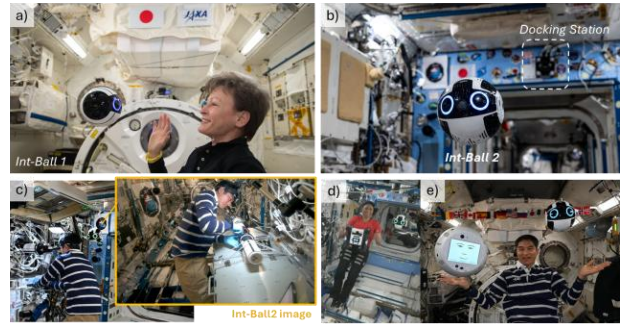

Fig. 4: JAXA's Int-Ball in "Kibo". a) First version of Int-Ball, b) Int-Ball2 with increased autonomy, c) monitoring astronaut's task with Int-Ball2, d) Int-Ball2 collaborating with Astrobee during Kibo-RPC, e) Int-Ball2 collaborating with CIMON in ICHIBAN mission. Image credit: JAXA/NASA

CIMON's AI is based on IBM's Watson platform, which runs on Earth servers. Spoken commands are transmitted to the ground, processed by Watson, and responses are sent back to the robot. Onboard software handles autonomous navigation and flight control via visual odometry and motion planning. CIMON is not self-learning; its knowledge base must be pre-loaded and trained by human operators. This cloud-dependent architecture contrasts with systems designed for onboard autonomy and was chosen for demonstration purposes.

CIMON has supported research in human-robot interaction, AI-based assistance, and space psychology. Its operational history includes two hardware versions and interactions with multiple ESA astronauts. The first version, CIMON-1, conducted its initial 90-minute session with astronaut Alexander Gerst in November 2018. During these and subsequent sessions, it successfully demonstrated voice-controlled navigation to specific points, displayed procedures for experiments, played music, and showcased its conversational capabilities. Based on CIMON-1's experience, CIMON-2 launched in 2019 with improved microphones, software stability, advanced AI, and 1h longer battery life. It assisted astronauts like Luca Parmitano in tasks such as guiding a water-crystal experiment and acting as an intelligent camera. CIMON also collaborated with astronauts on multi-step tasks, such as solving a Rubik's cube, showcasing its ability to follow complex interactions.

The primary lesson learned from the CIMON program is that on the one hand, the technologies provided and used by CIMON demonstrated that human-machine interaction with a verbal and visual interface is a suitable additional tool for the daily work in the human exploration environment. Most of the users have provided positive feedback. On the other hand, the current CIMON ecosystem is limited by the setup and dependencies from various services. The limitation is linked to the mission goal defined in 2016 of being an early available technology demonstrator. To allow the user to make use of the full performance of the CIMON idea, the next step would be a hardware update to increase autonomy and easy availability of the system (like having a docking station available and the free flyer permanently deployed) as well as on-premises deployed AI services based on Large Language Models and agentic AI.

More detailed lessons learned can be identified by deeper investigation and analysis of the user debriefing given so far and to be collected during its missions. The CIMON project is very fortunate in having users and science subjects interacting with the system since 2018 on-board the ISS, coming from over five nations and performing different and comparable tasks. Together with other stakeholders, new and precise system requirements can be identified which will guarantee a user-centered design evolution.

### C. Int-Ball (JAXA): Dedicated Camera Drone

JAXA has been pursuing crew-assistive robotic technologies. To develop, verify, and accelerate integration, JAXA uses the ISS to deploy, demonstrate, and gain operational experience. The JEM includes the Remote Manipulator System (JEMRMS), a remote-controlled robotic arm that reduces risk and optimizes Extra-Vehicular Activities (EVAs) [17]. Hardware automation, such as remote airlock control, has also been integrated into daily operations. JAXA has also conducted ISS demonstrations such as Rex-J (Robot Experiment on JEM) [18].

Building on this experience, JAXA established the Int-Ball missions to reduce crew workload during Intra-Vehicular Activities (IVA), especially for video-taking and monitoring. Analysis shows astronauts spend over 10% of their time on imagery tasks [19], which are crucial for effective ground-space collaboration. Int-Ball aims to shift the responsibility of camera setup and field-of-view adjustments to ground control, freeing up valuable crew time.

The first Int-Ball (JEM Internal Ball Camera Robot) was developed and launched to the ISS in 2017 as a technological demonstration [20] (Fig. 4a). It is a 15cm diameter spherical free-flying robot deployed in the "Kibo" module, controlled from the ground using its 12 fans (about 3mN each) and 3 reaction wheels for movement, and transmits images. It was tested on the ISS in 2017 and 2018, providing valuable insights into operational challenges. While its compact size and single-task focus suited crew interaction, its propulsion struggled with airflow disturbances from ISS ventilation, especially near ducts. It also required crew for deployment and recharging and relied on stereoscopic target markers for localization, which could be obscured by crew or objects.

To address these issues, JAXA developed Int-Ball2 [21] (Fig. 4b) with enhanced propulsion capabilities using 8 propellant fans (increased thrust force to 60-163mN) [22], self-deployment and recharging via a docking station [23], and improved localization using stereo-camera-based Visual Simultaneous Localization and Mapping (V-SLAM) and IMU fusion [21]. The ROS-based modular software, running on Jetson TX2 computer with Ubuntu, allows extensibility and easy software updates. It retains a spherical shape but increased to 20 cm in diameter to accommodate upgrades. The original spherical design was praised for robustness and omni-directional movement but lacked clear orientation. Int-Ball2's design now visually indicates direction for better ground communication. Int-Ball2 system is mainly composed of Commercial-Off-The-Shelf (COTS) components.

Ground testing combined software simulations (ROS/Gazebo with custom plugin for airflow simulations), hardware-in-the-loop (HIL) simulations, and hardware testing. To correlate the software simulations, force-torque sensor measurements of the propulsion force are reflected in the software simulations [24]. For hardware testing, the propulsion system employed two-dimensional granite floating tables. While three-dimensional movements cannot be imitated in such a setup, the correctness of the two translational directions and one rotational movement was tested and repeated for all axes prior to launch. To cover the disadvantage of the fully software-based simulation and the planar air-bearing platform testing, JAXA developed a HIL simulator simultaneously verifying the GNC systems in three dimensions, including flight hardware and software [25].

Int-Ball2 launched in June 2023. During its checkout, power, communication, imagery, flight, and automated charging were validated. Crew-interactive features and off-nominal transitions (e.g., low battery return) were tested. Flight data was logged, and additional reference markers were installed in JEM for navigation assessment [26]. The operational demonstration showed its capabilities in actual tasks, including automated release, flight in front of the airlock for crew interviews, module flight for consolidation tasks, and monitoring experiments with astronauts.

Following the successful checkout, Int-Ball2 became part of the JEM infrastructure. In 2024, it served as a camera robot to film Kibo-RPC's Astrobee work (Fig. 4d). Since 2025, it has been used for scientific tasks, including the photographing crew maintenance of the Electrostatic Levitation Furnace (ELF, shown in Fig. 4c), exchanging samples for the Protein Crystal Growth (PCG) experiment. Int-Ball2 also serves as a remote monitoring platform, reducing crew workload. It performed remote check of the fire indicator in JEM, a task which was previously performed annually by crew.

Its ROS-based modular software supports updates and custom integrations. Safety is ensured through hardware-based controls, isolating software changes. Int-Ball2 user programming platform is available open-source [27]. JAXA has tested this feature for complex trajectories, docking robustness, and vision-based object recognition. Future upgrades include autonomous path planning leveraging V-SLAM maps for obstacle avoidance and tracking [28]. In 2025 Int-Ball2 also collaborated with CIMON for real-time communication and data exchange (Fig. 4e) [29]. Using the same ROS middleware enabled such collaborations, presenting potential use-cases for data sharing and distributed task responsibilities.

## III. Lessons Learned

Development and operations of Astrobee, CIMON and Int-Ball have produced a convergence of experiences. This section synthesizes the lessons learned across these robots, organized by their development, operations, and user integrations.

### A. Development

*Importance of Early Prototyping:*

All three platforms underscore the value of iterative development. Early prototyping provided essential experience that informed relevant and verifiable requirements before formal reviews. For CIMON, building a prototype before the official design review accelerated development. Int-Ball1's quick prototyping enabled fast initial progress, though limited ground testing may have caused on-orbit issues—such as poor performance in strong airflow disturbances. Int-Ball2 addressed this by combining physical prototyping with software simulations, including Monte Carlo simulations and thrust force measurements, to ensure robustness.

*The Value of Operational Feedback*

Beyond initial prototypes, learning from prior on-orbit missions proved invaluable. The Int-Ball program represents a clear case of generational evolution, where direct reflection on Int-Ball1's on-orbit challenges was crucial for defining the requirements for Int-Ball2. Similarly, the Astrobee program built upon more than a decade of operational takeaways from its predecessor, the SPHERES facility. The CIMON project also had the opportunity to implement improvements from CIMON-1 to CIMON-2.

*Balancing Size, Capability, and Function*

A central design trade-off is balancing the robot's capability within physical sizing and resource constraints. For CIMON, the philosophy was that design followed function; its diameter was determined by the need to display a life-sized face on its screen, while its agility was constrained by safety requirements. The Int-Ball1 in contrast was driven by size definition first; the original's size was designed to be graspable in one hand. To achieve greater autonomy, Int-Ball2 increased in size, requiring a compromise on the one-handed graspability, but maintaining a small size. Its compact size received favorable feedback from the crew, reinforcing that non-intrusiveness is crucial for working in proximity with crew.

*Modular Design and potential for onboard updates*

A modular design was a common strategy for managing complexity. Astrobee, CIMON, and Int-Ball2 all implemented modular software architecture including ROS middleware. While each project puts strong efforts on extendibility of the robot through modularity, physical modularity is a trade with size and resource limitations. Astrobee design choices improved modularity by decoupling propulsion from core avionics, isolating robust real-time code from experimental code by distributing computing across 3 processors, and employing a modular flight software architecture. However, unforeseen interface issues, like the Electromagnetic Interference (EMI) from propulsion modules injected onto the main power bus and the nozzles' acoustic noise amplified by the plenum, exposed limitations, and led to project delays. Int-Ball2 in contrast deliberately sacrificed some of the hardware replicability (limiting hardware modularity to propeller modules[21]) to achieve its compact body assembly.

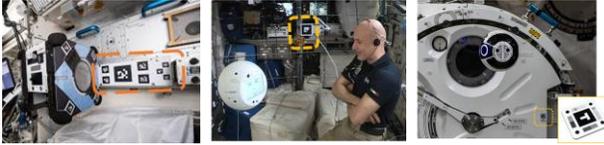
Fig. 5: Example of the different visual fiducials used by robots. Image credits: NASA/ESA/JAXA

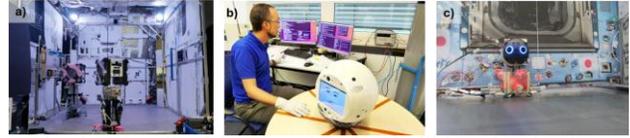
Fig. 6: Example of the ground testing facilities of each robot. Image credits: NASA/DLR/Airbus/JAXA

### *Strength and challenges of COTS*

These robots' designs applied many COTS components decreasing their costs and accelerating the development based on the ground robotics research heritages. However, this also introduced challenges. The Int-Ball team found that COTS specifications did not always match real-world performance and that material selection is complicated by stringent safety requirements, like non-flammability. As an example, Astrobee failure of commercial-grade SD cards in both the robots and the docking station led to multiple on-orbit repairs.

### *Leveraging Common Resources*

The robots described leverage the existing infrastructure of the ISS (e.g. power supply, heat-rejection, communications). Commonality in its components can also be a strong advantage. The CIMON project was also supported by Astrobee Team by making batteries available to the project in-orbit until CIMON received its own items. A common pool of e.g. batteries, chargers, navigation markers, and on-board computational power could accelerate the development and optimize operation resources. Visual markers used in each project were independent. Fig. 5 shows fiducials used by Astrobee for docking station, CIMON for absolute localization reference, Int-Ball2 for docking and for the independent navigation assessment. Interoperability of such interface could simplify each system and give potential to assess each robot's capabilities with same metrics data. Software interoperability with ROS is anticipated to enable sharing of sensory or computational resources, presenting promising path in resource constrained applications [29].

### *Incorporating Safety from the Initial Design Phase*

It is crucial to reflect human safety considerations, such as battery safety and collision avoidance, as well as compliance with ISS standards. A key lesson for enabling safe and rapid software updates is to anchor safety in hardware. The Int-Ball2 team assured its safety primarily through its hardware design (e.g., a low-impact shell, propulsion limits), not through software controls. During the development, Int-Ball2 face challenges with finding small-sized batteries that could fit-in to the assumed size of the robot while assuring all safety controls required – finally requiring dedicated development of battery units. Uniqueness of the free-flying robot also requires specific considerations for safety controls. Hazard control on the potential impact with crew or station equipment is incorporated in these free-flying robots. In such a process, CIMON project has shown that early alignment with Human Safety board supported smooth development phases and control of all hazards were crucial.

### *Verifying Microgravity Systems on Earth*

Testing a robot designed for microgravity is a fundamental challenge. All teams used a combination of simulation and hardware tests. However, some performance variances are only discovered on-orbit. To mitigate this, the Astrobee project maintains the Granite Testing Facility(Fig. 6a), which includes two ground units (*Bsharp* and *Wannabee*) that are identical in hardware and software to the flight units. This enables high-fidelity ground testing and anomaly replication. Int-Ball2 also uses granite table for hardware verification (Fig. 6c), combined with software simulations to verify three-dimensional movements [24, 25].

## B. Operation

### *Importance of Operational Autonomy*

Astrobee and Int-Ball2 experience shows that for frequent operations, operational autonomy is a key. While SPHERES, Int-Ball1, CIMON required crew for the setup of the robot, Astrobee and Int-Ball2 employed a dedicated docking station enabling more autonomous operation of the robot. Although such needs depend on the objective of the system, considering the limited crew resources, operational autonomy can significantly ease operations.

### *Redundancy in Communication*

Reliable communication is the bedrock of remote robotic operations. The first Int-Ball relied only on wireless communication, which caused loss of communication functions during its operations. While Int-Ball2 added a wired UART communication option when docked, on-orbit Wi-Fi performance can still degrade real-time video. Need for redundant communication was also highlighted during two separate Astrobee dock anomalies where the Ethernet connection failed; because the robots could still receive commands via Wi-Fi, operations continued with minimal disruption. For CIMON, the need for a continuous link for its cloud-based AI is a core operational constraint.

### *Navigation in the ISS Environment*

Localization is one of the most fundamental capabilities of a mobile robot, and the dynamic ISS environment poses a persistent challenge. Although it is a well-known environment, ISS internal setup changes due to crew work. During early Astrobee activities, its usual localization fix was occasionally lost, requiring operators to recover using lower-level teleoperated motion commands or even crew to help return Astrobee to its dock. This problem was mitigated at first by improving map imagery coverage, updating maps frequently to reflect environment changes, and planning activities around the need to occasionally renew the localization fix by visiting areas more favorable to map recognition. Later improvements to Astrobee's localization algorithms made it more robust to incomplete or out-of-date maps and relaxed some of these concerns [6, 30, 31, 32]. Int-Ball2 used a stereo-camera based visual-SLAM adapting to environmental change. However, a relative pose estimation has some shifts with cumulative errors. Post-flight analysis shows that about 8% errors are encountered during the flight causing a localization shift for more complex, longer maneuvers. While Int-Ball2 tackles this issue with planning a path for better loop-closures, addition of the absolute navigation references could reduce such errors.

### On-Orbit Anomaly Resolution and Maintenance

A key operational lesson is the necessity of a structured anomaly resolution strategy. The process involves attempting to reproduce faults on identical ground hardware, followed by remote recovery attempts (e.g., power cycles, software patches like watchdogs). Such steps have been taken by all teams. If remote recovery fails, the next step is on-orbit hardware replacement by the crew, such as replacing failed SD cards in the Astrobee docking station. The final option is a hardware replacement or to down-mass the faulty hardware for ground repair. Int-Ball and CIMON projects replaced the robots to upgraded version, Astrobee "Honey" was brought to ground for the repairs. Such a tiered approach is essential for supporting long-term operations.

### Crew Privacy

Operating alongside astronauts necessitates a focus on the human element. A key consideration is crew privacy. CIMON provides a mechanical "offline" button, while Int-Ball2 indicates data acquisition with LEDs. Astrobee's design integrated a series of visual cues that helped crew understand that the robot was operational and therefore video may be streamed down to ground. Prior to all the robots' activities, crew were informed for their awareness.

### Reflection of Ground Operations Loads

Operational experience outlines the need for reflection on ground operators' fatigue as well. For Int-Ball2, ground controllers are required to plan its behaviors and movement path in real-time during operations. Although it uses graphical user interface to control the robot, the simple commanding and checking its successful implementation often needs 1-3 minutes for execution due to the procedural rules. This poses a challenge in real-time collaborative work with astronauts. Astrobee's operations relies on the effectiveness of the Ground Data System's (GDS) Astrobee Control Station and GDS Helper (GDS-H) graphical user interfaces (GUIs), which are used by ground operators to monitor and control the robots in real-time. The Astrobee facility carried out a study to better understand the impact GDS has in the operator's cognitive load during an ISS activity; GDS Helper serves as a comparison baseline in this study. The study findings recommended a hybrid approach combining the single-window simplicity of GDS-H with the situational awareness tools of GDS, potentially reducing cognitive load while maintaining effective control of the robots [33].

### Data Management

For high-volume data producers like Astrobee, collecting over 2.0 Terabytes of data forced the development of new, streamlined processes for data clearance and distribution e.g. storing data at the dock at ISS, and a data web server on the ground where Guest Scientists may access it. Similarly, Int-Ball2 with its increased operation faces challenges for ground operator work to downlink required data. To minimize such effort, although Int-Ball2 can acquire 4K imagery, its operators favor usage of Full-HD image whenever higher resolution is not necessary.

## C. User Integration

### Managing a Multi-User Robotics Facility

Operating a facility for a wide range of users presents unique challenges. The Astrobee facility has successfully supported over 175 on-orbit sessions, demonstrating a high operational tempo. However, frequent anomalies often led guest scientists to rely heavily on the Astrobee operations team for real-time command and control. This highlights the need for robust user support, high-fidelity ground test environments, and clear procedures for user-led software development to ensure guest scientist success.

### The Importance of a User Community

A major factor in Astrobee's success as a research platform is its open-source software and the cultivation of a user community. An outstanding example of international collaboration enabled by the Astrobee community was showcased by Canadian Obruta using the hardware payload from Australian CSIRO's Multi-Resolution Scanner. Obruta ran their computer-vision algorithms on CSIRO's payload and demonstrated their proposed solution worked. The usage of ROS in all these robots also lowers the barrier to entry for guest scientists and leverages existing research. It also enables powerful collaborations, such as JAXA's Kibo-RPC, which uses Astrobee as an educational platform for students worldwide, or ICHIBAN where Int-Ball2 and CIMON collaborated in real-time. The wide user community of ROS enables easier implementation by the users. However, for more safety-critical software implementation more reliable middleware framework might be desirable.

### Unused Potential

The advanced HRI elements on Astrobee, such as its laser pointer and signal lights, have been used effectively but only by a small number of guest scientists specifically studying HRI or needing to convey robot state, such as in the Kibo-RPC. This experience suggests that unless a user's research is specifically focused on HRI, they are unlikely to incorporate these features, relying instead on core mobility and sensing capabilities. CIMON includes hardware like a laser pointer and USB ports that can be activated in the future. Int-Ball2 also has not used its USB payload interface (as of Oct. 2025).

## IV. CONCLUSION AND PROSPECTS

The experience of free-flying robots on the ISS offers a comprehensive view of the state-of-the-art in intra-vehicular robotic astronaut assistance and technology demonstrations. This paper presented joint lessons learned from development, operations, and user integration.

In the development phase, complex tradeoffs in the resource constrained conditions were carried out with rapid prototyping. While each project had distinct objectives, shared experience emphasized the importance of adaptability to ISS infrastructure and design commonality. Software interoperability via mutual middleware architecture (ROS) enables promising multi-robot collaboration, especially in resource-constrained environments. In these aspects, efforts like Space-ROS [34] development are highly anticipated. Operationally, navigating a dynamic, human-centered environment proved challenging. These experiences underscore the need for robust localization research and operational strategies to mitigate constraints, as well as opportunities to improve absolute localization in future modules. On-orbit operations revealed "hidden workloads" for astronauts and ground controllers. Variable lighting, airflow disturbances, and changing configurations posed persistent challenges to autonomy. Operational costs, including ground operator workload, crew time for setup/stowage, and environmental preparation, often limited robot utility. A key lesson is balancing autonomy with operational predictability to build crew trust. Robot effectiveness depends on resilience (hardware/software), ground operability and maintainability, and the social dynamics of crew interaction.

Looking ahead, these lessons point to a future focused on integration and enhanced autonomy. Architecturally, this includes robust onboard edge AI and hybrid navigation systems. Programmatically, iterative design cycles and minimizing operator cognitive load will be essential for sustainable ground operations. Future platforms should balance interoperability and standardization with reliable core systems for ease of use, while offering extensibility for advanced research partners to push scientific boundaries. In conclusion, the knowledge shared here is expected to support the future development of crew-assistive robots, serving in the enhancement of human space exploration.

ACKNOWLEDGMENTS

The work presented is a result of the collective effort of each team's members, including many partners and contractors who supported development, operations, and user integration for Astrobee, CIMON, and Int-Ball.